\documentclass{esannV2}
\pdfoutput=1
\usepackage{graphicx} 
\usepackage[latin1]{inputenc}
\usepackage{amssymb,amsmath,array}
\usepackage{float}
\usepackage{amsfonts}
\usepackage{theorem}
\usepackage{amscd}
\usepackage{latexsym}
\usepackage{dsfont}
\usepackage[english]{babel}
\usepackage{url}
\usepackage{microtype}

\DeclareMathOperator*{\argmin}{argmin}

\newcommand{\zbf}{{\bf z}}

\newcommand{\real}{\mathbb{R}}

\begin{document}
\title{A Discussion on Parallelization Schemes for Stochastic Vector Quantization Algorithms}

\author{Matthieu Durut$^{1,3}$, Benoit Patra$^{2,3}$, Fabrice Rossi$^{4}$
\vspace{.3cm}\\
1- Telecom ParisTech - INFRES, 46 rue Barrault - Paris - France
\vspace{.1cm}\\
2- Université Pierre et Marie Curie - LSTA, 4 place Jussieu - Paris - France
\vspace{.1cm}\\
3- Lokad , 10 rue Philippe de Champaigne - Paris - France
\vspace{.1cm}\\
4- SAMM, Université Paris I Panthéon-Sorbonne - 90 rue de Tolbiac - Paris - France
}

\maketitle

\begin{abstract}
This paper studies parallelization schemes for stochastic Vector Quantization
algorithms in order to obtain time speed-ups using distributed resources. We
show that the most intuitive parallelization scheme does not lead to better
performances than the sequential algorithm. Another distributed scheme is
therefore introduced which obtains the expected speed-ups. Then, it is
improved to fit implementation on distributed architectures where
communications are slow and inter-machines synchronization too costly. The
schemes are tested with simulated distributed architectures and, for the last
one, with Microsoft Windows Azure platform obtaining speed-ups up to $32$
Virtual Machines.
\end{abstract}

\section{Introduction}
Motivated by the problem of executing clustering algorithms on very large
datasets, we investigate parallelization schemes of the stochastic Vector
Quantization (VQ) method (also called online $k$-means). This procedure is
known for its good statistical properties but it does not exhibit the
embarrassing parallelism of the (batch) $k$-means. Given a satisfactory
sequential implementation of the VQ algorithm, we aim at speeding up its
execution through a parallel implementation: the ultimate goal is to reduce
the wall time used by the method on a given dataset, that is the time needed
to reach some performance threshold, using more than one computing
unit. Theoretical parallel VQ algorithms are studied in \cite{PAT1}. The aim
of the present paper is to derive actual real world implementations.

The VQ technique computes a summary of a dataset $\{\zbf_{t}\}_{t=1}^n$ of $d$
dimensional samples with $\kappa$ prototypes, $w=(w_1, \ldots,w_\kappa) \in
\left(\real^d\right)^\kappa$. Starting from a random initial $w(0) \in \left(\real^d\right)^\kappa$
and given a series of positive \emph{steps} $(\varepsilon_t)_{t>0}$, VQ
produces a series of $w(t)$ by updating $w$ prototype by prototype. More
precisely, with $l(t)=\argmin_{i = 1, \ldots, \kappa} \left\|\zbf_{\{t+1
    \text{ mod } n\}}  - w_i(t)\right\|^2$, we have
\begin{equation}\label{eq:finiteSample}
w(t+1)_i=
\begin{cases}
  w(t)_i&\text{when $i\neq l(t)$}\\
w(t)_i-\varepsilon_{t+1}(w(t)_i-\zbf_{\{t+1    \text{ mod } n\}})&\text{when $i=l(t)$}\\
\end{cases},
\end{equation}
where the mod operator stands for the remainder of an integer division
operation. A theorem about almost sure convergence of the VQ procedure is proved in
\cite{PAG1}. It is well known that the VQ algorithm belongs to the class
of stochastic gradient descent algorithms (see  \cite{PAG1} for instance).

This paper follows the VQ ideas presented in \cite{PAT1}. We assume having access to $M$ computing entities, each of them executing concurrent VQ procedures. These executions are performed on a dataset, split among the local memory of the computing instances, and represented by the sequences $\{\zbf_t^i\}_{t=1}^n$, $i\in\{1,\ldots,M\}$. The prototype iterations computed by the VQ techniques on each node are denoted by $\{w^{i}(t)\}_{t=0}^\infty$ and called \emph{versions}. We use the following normalized criterion to measure the speed-up ability of our investigated schemes.
\begin{equation}\label{eq:empiricalDisto}
C_{n,M}(w) = \frac{1}{nM} \sum_{i=1}^M \sum_{t=1}^n{\min_{\ell = 1, \ldots, \kappa} \left\|\zbf^i_t  - w_\ell\right\|^2}, \quad \mbox{$w \in \left(\real^d\right)^\kappa$}.
\end{equation}

The rest of the paper is organized as follows. First, Section \ref{sec:firstScheme} provides
empirical evidences that the most simple scheme cannot bring speed-ups. Then,
some insights to explain the previous non satisfactory situation are provided
in Section \ref{sec:secondScheme}. Consequently, we design a new scheme and
prove by practice its ability to bring speed-ups. Finally, in Section
\ref{sec:delaysScheme}, we present an asynchronous adaptation of this latter
scheme which fits better slow communication architectures such as Cloud
Computing.

Notice that the proposed algorithms are tested using simulated distributed
architecture and synthetic vector data\footnote{The source code is available
  at \url{http://code.google.com/p/clouddalvq/}. Details about the artificial
  data generator are available in Section 4.2 of
  \url{http://www.lsta.upmc.fr/doct/patra/publications/PhDMain.pdf}.}, but our conclusions are more sensitive to the loss function smoothness and convexity than to the data choice.
  

\section{A first distributed scheme}\label{sec:firstScheme}
\begin{figure}[h]
\begin{center}
\includegraphics[scale = 0.35]{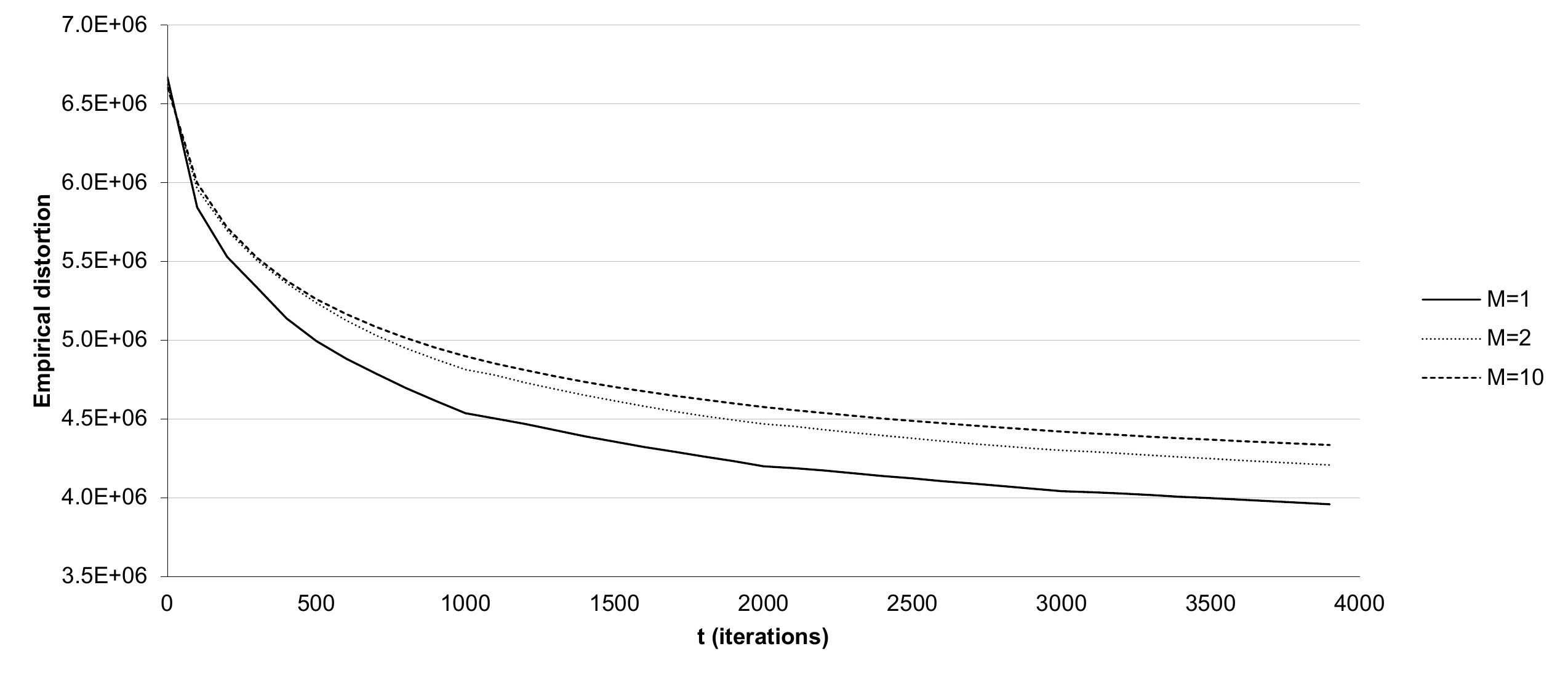}
\end{center}
\caption{Charts of performance curves for iterations (\ref{eq:averaging}) with $\tau=10$ and different number
of computing entities: M = 1, 2, 10.}
\label{fig:parallel10}
\end{figure}

Our investigation starts with the most intuitive parallelization scheme. Each
computing resource starts with the same initial prototypes (a.k.a., versions):
$w^1(0) = \ldots =w^M(0)$. Then each machine applies the sequential VQ to its
subset of the dataset. Once in a while, prototypes are synchronized: when
$\tau$ data points have been processed by each concurrent processor, a shared
version of the prototypes is computed as follows (here for the first
synchronization event):
\begin{equation}\label{eq:averaging}
w^{srd} =  \frac{1}{M}\sum_{j=1}^M{w^j}(\tau).
\end{equation}
The shared version is then broadcasted to each processing unit. In the case of a smooth convex loss function, distributed stochastic gradient descent algorithms with averaging of local results provide a speed-up in comparison of the sequential algorithm (see \cite{OptimalOnlinePrediction}). Figure \ref{fig:parallel10}
shows a typical evolution through wall time of the quantization error, obtained with an execution of the scheme on a simulated parallel implementation
in which communications are instantaneous. It shows up that in our non smooth and convex loss function case, multiple resources do
not bring speed-ups for convergence. Even if more data are processed, no gain
in term of wall clock time is provided using this parallel scheme.

\section{Towards a better scheme}\label{sec:secondScheme}
The investigation of the previous non-satisfactory result starts by rewriting
both the sequential and the distributed scheme. Let us first
introduce $H(\zbf,w)$ defined by
\begin{equation}\label{def:defH}
H(\zbf,w) = \left( \left(w_\ell - \zbf\right) \mathds{1}_{\left\{l= \argmin_{i = 1, \ldots, \kappa} \left\|\zbf  - w_i\right\|^2\right\}}\right)_{1 \leq \ell \leq \kappa}.
\end{equation}
Then, a series of the sequential VQ iterations (\ref{eq:finiteSample}) can be
rewritten:
\begin{equation}\label{eq:iteration}
w(t+1) = w(t - \tau + 1)- \sum_{t'=t-\tau+1}^{t} \varepsilon_{t'+1}H\left(\zbf_{\{t'+1\text{ mod }n \}},w(t')\right),\quad \mbox{$t \geq \tau$}.
\end{equation}
Then, just after a synchronization (defined by $t \text{ mod } \tau =0$ and $t
> 0$), for all $i\in\{1, \ldots,M\}$, the sequential VQ iterations on each
computational resource can be rewritten as follows
\begin{equation}\label{eq:iteration2}
w^i(t+1) = w^i(t - \tau + 1) - \sum_{t'= t- \tau +1}^{t}\varepsilon_{t'+1}\left(\frac{1}{M}\sum_{j=1}^M H\left(\zbf^j_{\{t'+1\text{ mod }n\}},w^j(t')\right)\right).
\end{equation}
Assuming that $w^j(t')\thickapprox w^i(t')$, for all $(i,j) \in
\{1,\ldots,M\}^2$ and $t' \geq 0$, the mean in parenthesis is an estimator of
the gradient of the distortion at $w^i(t')$. Consequently, the two algorithms induced by iterations (\ref{eq:iteration}) and (\ref{eq:iteration2})
can be thought as stochastic gradient descent procedures with different
estimators of the gradient but driven by the same learning rate which is given
by the sequence $(\varepsilon_t)_{t>0}$.

The convergence speed of a non-fixed step gradient descent procedure is
essentially driven by the decreasing speed of the sequence of steps
. The choice of this sequence is subject to an
exploration/convergence trade-off. Since the two procedures above share the
same learning rate with respect to the iterations $t \geq 0$, they share the
same convergence speed with respect to the wall clock time (time measured by
an exterior observer). Yet, the distributed scheme of Section
\ref{sec:firstScheme} has a much lower learning rate with respect to the
number of samples processed, favoring exploration to the detriment of the
convergence. The multiple resources therefore lead to better exploration but
to similar convergence speed with respect to wall clock time.

\begin{figure}[h]
\begin{center}
\includegraphics[scale = 0.35]{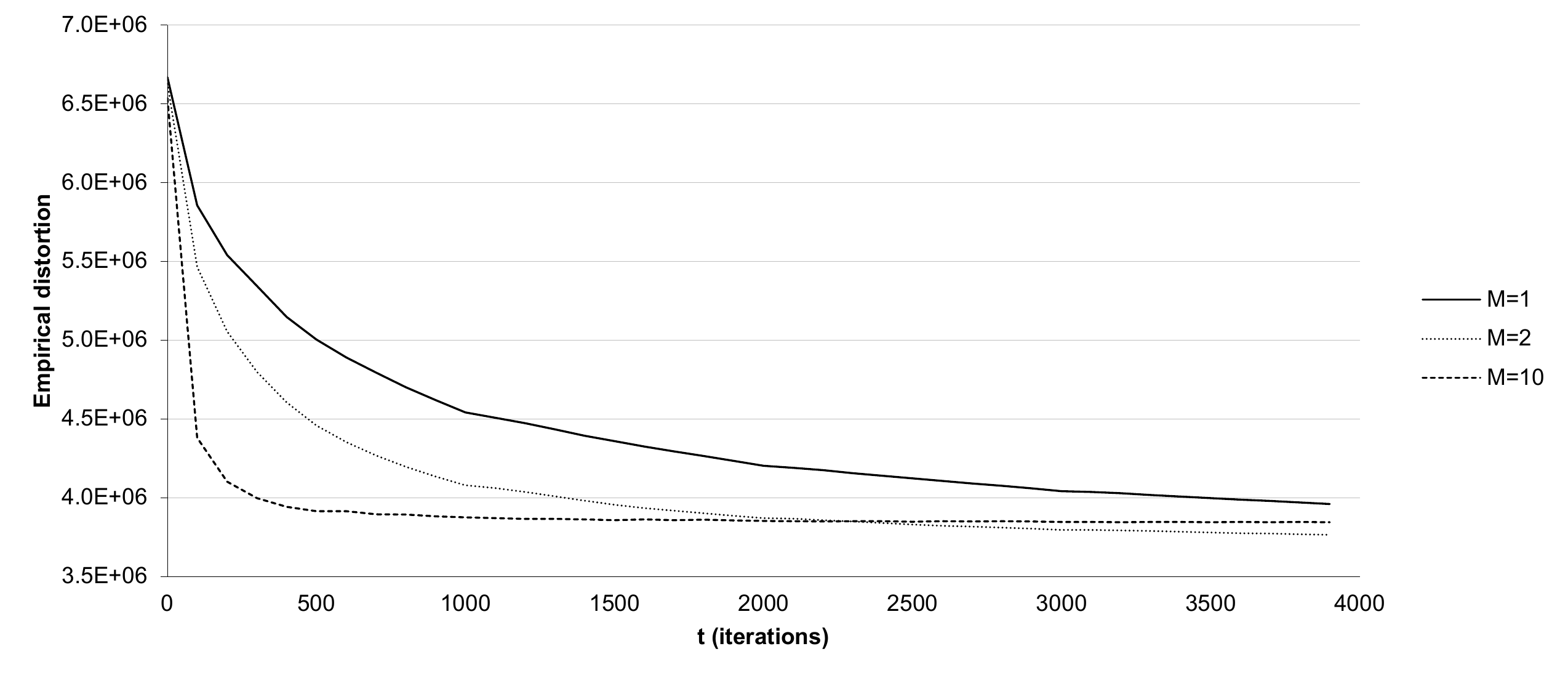}
\end{center}
\caption{Charts of performance curves for iterations (\ref{eq:correctAveraging}) with $\tau=10$ and different number
of computing entities: M = 1, 2, 10.}
\label{fig:correctedParallel}
\end{figure}

As we assume to have a satisfactory VQ implementation, the series of steps
$(\varepsilon_t)_{t>0}$ is supposed to be adapted to the dataset. Consequently
we should seek for a distributed scheme that have the same learning rate
evolution in term of processed samples and which convergence speed with
respect to iterations is accelerated. Denoting
\begin{equation}\label{eq:displacement}
\Delta^j_{t_1\rightarrow t_2} = \sum_{t'=t_1 + 1}^{t_2} \varepsilon_{t'+1}H\left(\zbf^j_{\{t'+1\text{ mod }n \}},w^j(t')\right), \mbox{ $j \in \{1, \ldots, M\}$ and $t_2 > t_1 \geq 0$}.
\end{equation}
At time $t=0$, $w^1(0) = \ldots =w^M(0)=w^{srd}$. For all $i \in \{1, \ldots,M\}$ and all $t \geq0$, consider the distributed scheme given by
\begin{equation}\label{eq:correctAveraging}
\begin{cases}
w^i_{temp} = w^i(t) - \varepsilon_{t+1} H\left(\zbf^i_{\{t+1\text{ mod }n\}},w^i(t)\right)& \\  
w^i(t+1) = w^i_{temp} & \text{ if }  t \text{ mod } \tau  \neq 0 \text{ or } t=0,\\
    \begin{cases}
        w^{srd} = w^{srd} - \sum_{j=1}^M{\Delta^j_{t- \tau \rightarrow t} } & \\
        w^i(t+1) =  w^{srd} &\\
    \end{cases}& \text{ if } t \text{ mod } \tau =0 \text{ and } t \geq \tau.
\end{cases}
\end{equation}
The main difference between the two parallel schemes consists in the way
results are merged in the reducing phase (described by the braced inner
equations): here we apply the translation calculated by each parallel VQ to
the current shared version of the prototypes, rather than averaging this
translation. The results of a typical application of this scheme are displayed in the charts of Figure \ref{fig:correctedParallel}. The charts show that substantial speed-ups are obtained with distributed resources. The acceleration is greater when the reducing phase is frequent. Indeed, if $\tau$ is large then more autonomy has been granted to the concurrent executions, they could be attracted to different regions that would slow down the consensus and the convergence.

\section{A model with stochastic delays}
\label{sec:delaysScheme}
The previous parallelization schemes do not deal with communication costs
introduced by update exchanges between machines. In the context of cloud
computing, no efficient shared memory is available and these costs introduce
delays. The effect of delays for parallel stochastic gradient descent has
already been studied (see for instance \cite{LAN1}) but for a computing architecture endowed with an efficient shared memory. Moreover, the unreliability of the cloud computing hardware introduces strong straggler issues and makes the synchronization process inappropriate. In this subsection, we improve the model of iterations (\ref{eq:correctAveraging}) with random communication costs that follow a geometric distribution and we remove the synchronization process of reducing phase, resulting in the more realistic iterations (\ref{eq:delayedAveraging}) below. For each time $t \geq0$, let $\tau^{i}(t)$ be the latest time before t when the unit $i$ finished to send its updates and received the shared version. At time $t=0$ we have $w^1(0) = \ldots =w^M(0)=w^{srd}$, and for all $i \in \{1, \ldots,M\}$ and all $t \geq0$,

\begin{equation}\label{eq:delayedAveraging}
\begin{cases}
w^i_{temp} = w^i(t) - \varepsilon_{t+1} H\left(\zbf^i_{\{t+1\text{ mod }n\}},w^i(t)\right)& \\ 
w^i(t+1) = w^{i}_{temp} & \text{ if }  t \neq \tau^{i}(t) \\
w^i(t+1) = w^{srd}(\tau^{i}(t-1)) - \Delta^i_{\tau^{i}(t-1) \rightarrow t}& \text{ if }  t = \tau^{i}(t) \\
w^{srd}(t+1) = w^{srd}(t) - \displaystyle\sum_{j, t = \tau^{j}(t)}{\Delta^j_{\tau^{j}(\tau^{j}(t-1) -1) \rightarrow \tau^{j}(t-1)}}& \\
\end{cases}
\end{equation}

\begin{figure}[h]
\begin{center}
\includegraphics[scale = 0.35]{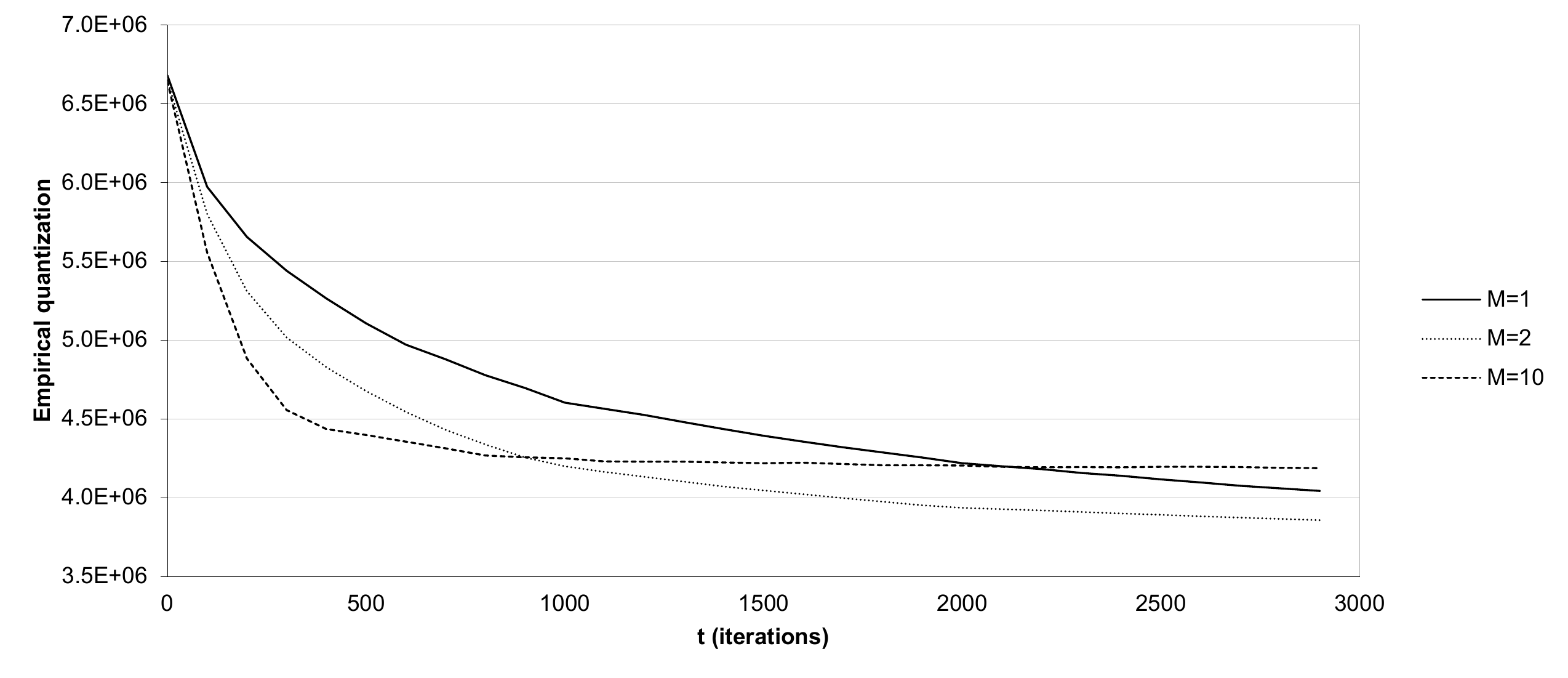}
\end{center}
\caption{Charts of performance curves for iterations (\ref{eq:delayedAveraging}) with $\tau=10$ and different number
of computing entities: M = 1, 2, 10.}
\label{fig:delayedParallel}
\end{figure}

\begin{figure}[h]
\begin{center}
\includegraphics[scale = 0.35]{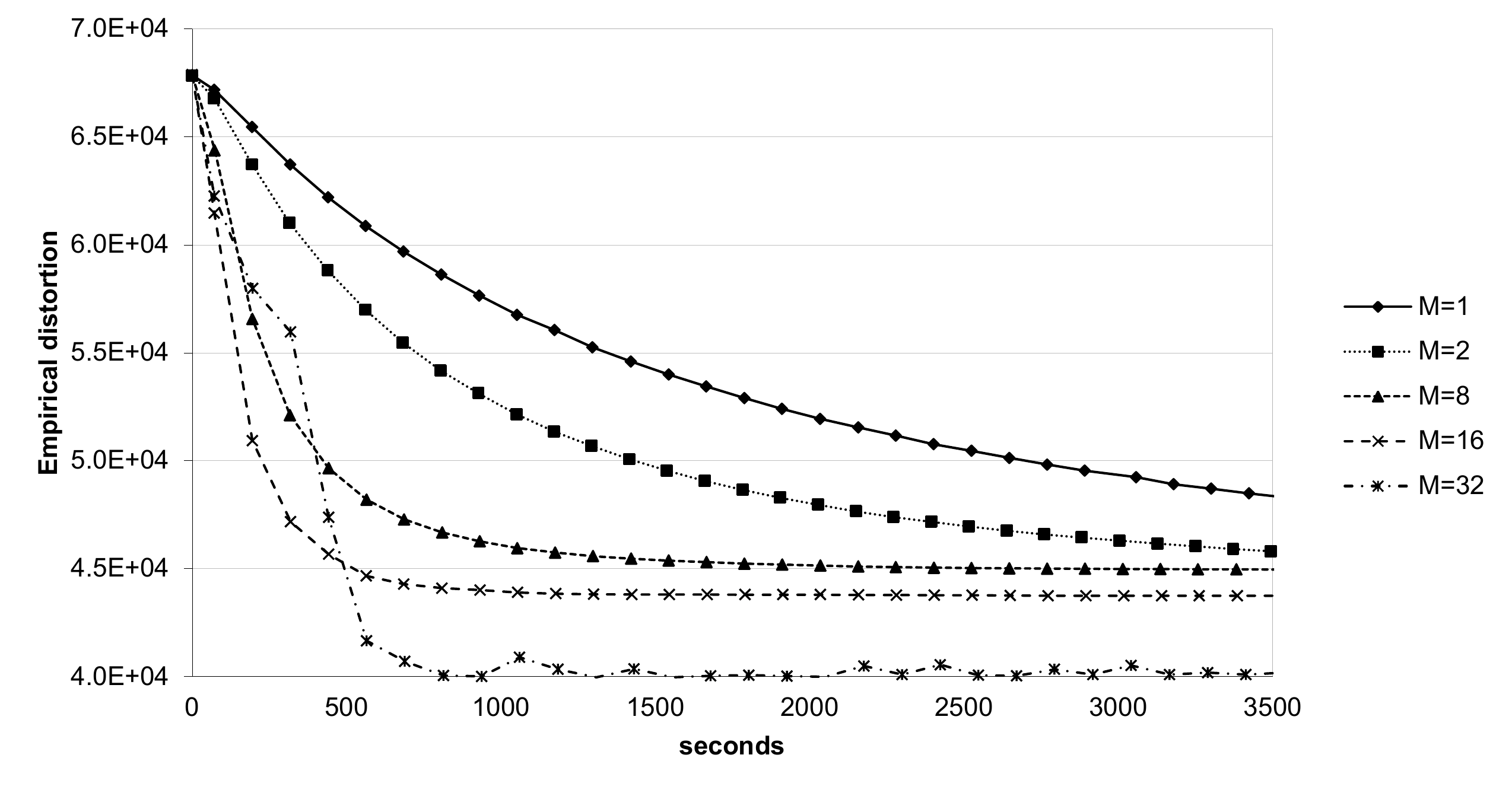}
\end{center}
\caption{Charts of performance curves for iterations (\ref{eq:delayedAveraging}) on our cloud implementation and different number
of computing entities.}
\label{fig:cloudParallel}
\end{figure}

There are no more synchronization between processing units: each machine uploads its updates and downloads the shared version as soon as its previous uploads and downloads are completed. A dedicated unit permanently modifies the shared version with the latest updates received from the other machines without any synchronization barrier. The Figure \ref{fig:delayedParallel} shows that the introduction of small delays and asynchronism only slightly impacts performances, compared to the scheme given by equations (\ref{eq:correctAveraging}). The Figure \ref{fig:cloudParallel} shows the results obtained by our cloud implementation\footnote{\texttt{http://code.google.com/p/clouddalvq/}} of the iterations (\ref{eq:delayedAveraging}) using 32 real processing units. A future paper will describe more precisely this cloud implementation.

\section{Conclusion}

In this paper we show that the naive parallelization scheme proposed in Section \ref{sec:firstScheme} does not provide better performance than the sequential scheme. This surprising result derives from the fact that our first parallel scheme leads to a decrease of the learning rate per data points processed. We therefore propose a new parallelization scheme relying on asynchronous updates of a common "shared version". This latter algorithm is very well suited for parallel computation on slow communication networks such as cloud computing platforms. Our implementation on Azure shows significant scale-up, up to 32 machines.

\begin{footnotesize}

\bibliographystyle{unsrt}
\bibliography{dalvq_esann_2012}

\end{footnotesize}


\end{document}